# Speed and Conversational Large Language Models (LLMs): Not All Is About Tokens per Second

Javier Conde, *ETSI de Telecomunicación, Universidad Politécnica de Madrid, 28040 Madrid, Spain*

Miguel González, *ETSI de Telecomunicación, Universidad Politécnica de Madrid, 28040 Madrid, Spain*

Pedro Reviriego, *ETSI de Telecomunicación, Universidad Politécnica de Madrid, 28040 Madrid, Spain*

Zhen Gao, *Tianjin University, 300072 Tianjin, China*

Shanshan Liu, *University of Electronic Science and Technology of China, 611731 Chengdu, China*

Fabrizio Lombardi, *Northeastern University, Boston, MA 02115, USA*

*Abstract*—The speed of open-weights Large Language Models and its dependency on the task at hand, when run on GPUs is studied to present a comparative analysis of the speed of the most popular open LLMs. The results suggest that existing token-based speed metrics do not necessarily correlate with the time needed to complete different tasks.

A few years ago, a Large Language Model (LLM) was un unknow entity. Today, following the introduction of ChatGPT[1] by OpenAI in November 2022, it is hard to find someone that does not know conversational LLMs as many have used them. The widespread adoption of LLMs has fostered the development of advanced models and tools based on these models, such as GPT4[2] by OpenAI, or Gemini[3] by Google and consisting of hundreds of billions of parameters. Unfortunately, these models are closed and can only be accessed through the user interfaces, tools or application programming interfaces provided by the companies that developed the models. Their parameters and implementation details are not publicly available and even if they were, their huge size would make their execution on commodity computing devices unfeasible. A different approach has been taken by some large companies such as Meta, i.e. the code as well as the parameters or weights of LLMs such as LLaMa[4] have been released for public use. This approach has been followed by startups such as Mistral releasing several open models, such as Mistral-7B[5], 01.AI with the Yi models and more recently Google with the Gemma models; all of these models have been publicly released.

These open LLMs provide all the parameters and code needed to run the models locally, but they typically do not disclose the data and procedures used for training and may place some restrictions on the use of the model, so they are not strictly speaking open-source. Therefore, next we refer to them as open-weights or simply open LLMs. In any case, open LLMs create new avenues for innovation and for democratizing access to LLMs.

Performance of LLMs in different tasks is thoroughly evaluated using benchmarks[6] that test their knowledge on a wide range of topics and their ability to perform many different tasks, like reasoning or problem solving in different areas. There are even public arenas[7] in which users can compare models and mark their preferences. These performance comparisons can guide users when selecting a model for a given task or application; however, performance on the task at hand is not the only metric of interest, cost and speed are also important.

In the case of closed commercial models, the cost is determined by their service rates typically charging per token; the speed depends on both their capacity to serve many clients and the limits of requests per unit of time set in their user agreements. For open models, the cost depends on whether the models are run locally or on the cloud. In the first case, there is a one-time investment on the computing infrastructure and operating expenses related for example to energy consumption or technical support. For cloud deployments, the cost is typically related to the amount of time that the allocated computing resources are used.

In both cases, LLMs' speed and memory usage are important factors because they determine the hardware requirements and to some extent the energy needed to support a given number of requests per second which, in turn impact cost for both cloud and local deployments. The memory needed by LLMs is roughly proportional to the number of parameters and the format used to represent them. However, the understanding of the speed of LLMs is more complex because it depends on different factors such as the target computing unit (i.e., typically a Graphics Processing Unit, GPU), the model architecture, the format of the parameters, and the number of requests that are processed at the same time. In this paper we pursue the evaluation of the speed of several open-weights LLMs of similar sizes when run on GPUs. The results suggest that existing token-based speed metrics do not necessarily correlate with the time needed to complete different tasks.

.

## THE OPEN-WEIGTHS LLM ECOSYSTEM

The availability of powerful open LLMs that can be modified, integrated with other applications and run locally, has spurred an ecosystem with hundreds of LLMs of different sizes, parameter formats, languages supported, and customization for a given task. These LLMs are readily available, and several tools and libraries have been developed to ease the execution of the models.

A common feature of most open-weights LLMs is that they can be run on a single computing unit, typically a GPU. This enables their use on commodity hardware such as computers with a GPU, or on single GPU instances, on the cloud. The main limitation to run on a single unit is the memory, low-end GPUs are typically equipped with a few GB of memory while high-end GPUs have tens of GB of memory. Each parameter of a LLM requires 2-4 bytes of memory when using traditional formats such as half and single precision floating-point representations[9]. Therefore, even when using half precision floating point, a 7 billion parameter model such as Mistral 7B requires approximately 14 GB of memory. This makes running larger models such as LlaMa-70B rather challenging, because 140 GB of memory is required, a memory size that is not available even in high-end GPUs. To address this issue, the open-source community has proposed alternative formats[10] for the parameters that require fewer bits, typically 8 or 4 and even close to 1 bit[11]. These formats enable for example running LlaMa-70B with 4 bits per parameter on a GPU that has only 40 GB of memory.

To evaluate the speed of executing LLMs, the first step is to select a subset of models. Evaluating all open models is unfeasible and most of them are fine-tuned versions of other models, so we expect them to have similar speed as the base model. In our investigation we focus on five models from four companies with similar sizes, around 7 billion parameters. In this case we can use the same format, 16-bit floating-point, for all of them so that comparisons are fair. The first two are LLaMa-2[4] and LLaMa-3 models from Meta with sizes 7B and 8B; a model from Mistral with 7B, another from 01.AI with 6B, and the last one from Google with 7B parameters (the models were taken from Huggingface, their exact names are "Llama-2-7b-chat-hf", "Meta-Llama-3-8B-Instruct", "Mistral-7B-Instruct-v0.1", "Yi-6B-Chat" and "gemma-7b-it", respectively). The models selected for evaluation are summarized in Table 1. This selection covers a wide range of models of a similar size that can be run on a single GPU; they are models that have been widely used as base to derive fine-tuned models. Therefore, the results obtained can be to some extent extrapolated to the derivative models.

TABLE 1. Models evaluated

| Model | Company | Sizes |
|---|---|---|
| LLaMa-2/3 | Meta | 7B/8B |
| Gemma | Google | 7B |
| Mistral | Mistral | 7B |
| Yi | 01.AI | 6B |

As said, for the parameter format, we have used floating-point formats with 16 bits for all models. This enables both running the models on a 40 GB memory and making a fair comparison among them. The results and insights obtained would be similar when using other formats, for example 8 or 4 bits.

Finally, in terms of computing platform, we consider the use of GPUs because they are widely used and accessible both locally and on the cloud. We used a high-end GPU from NVIDIA: the A100 with 40GB of memory which is enough memory to run all models in Table 1 as discussed before.

Modern GPUs have a large amount of processing units and when running a single prompt for a LLM, only a small fraction of those units is used. Since the parameters of the LLM are already in the GPU memory, they can be used to run several prompts at the same time, so creating batches of prompts that are run together. This can provide in many cases a significant speed up. The use of batches is possible when users run sets of prompts rather than individual prompts. For example, when creating questions for a test we can ask the LLMs to create questions on different topics, so using different prompts. Batches are also used when the GPU serves many users whose requests can be grouped. Therefore, we evaluate the models with different batch sizes to study the impact of batch size on the speed.

## EVALUATING LLM SPEED

The speed of LLMs is typically measured by the number of tokens generated per second, or the time needed to generate a given number of tokens, for example 256[8]. These metrics are easy to compute but they have some important limitations when we want to compare LLMs. The first limitation is that for two LLMs that have different tokenizers, it is not a fair comparison because the number of tokens for the same text are different. A second limitation is that even if two LLMs use the same tokenizer, then they may produce a different number of output tokens for the same task and thus, measuring the tokens per second does not fully capture the speed seen by the user. These limitations boil down to the fact that as users, we want to measure the speed of the model when performing a task, not the number of tokens generated.

To better understand the relative speed of LLMs, instead of measuring tokens, we measure the time needed to complete different tasks. First, we select 660 miscellaneous questions from a multiple-choice LLM benchmark[6] and use them for three tasks:

1) *Select the right choice for the multiple-choice questions*: the LLM only must generate the response with the selected choice (a, b, c, d). The model may produce additional text to explain its answer even though the prompt explicitly asks to answer only with the selected choice.
2) *Paraphrase the multiple-choice questions without the answers*: the LLM generates a text of similar size to the input text.
3) *Answer the questions providing an explanation for the answer selected*: the LLMs generate a textual answer with no constraints.

The overall approach is illustrated in Figure 1. The first task is designed to measure LLM speed when used to answer questions that require almost no text generation, just the option selected and a few additional words. The second task is designed to enable a comparison of LLMs when producing a similar amount of text as paraphrasing leaves little room for variations in text lengths. Finally, the third task is designed for LLMs to generate text freely to assess whether different LLMs generate texts of different lengths for the same prompts and its effect on speed.

The results for the first task are summarized in Figure 2. The plot on the left shows the time needed on average to generate a token (which corresponds to the process by which speed is commonly measured for LLMs), and in the middle, the time needed to answer all questions. The number of output tokens is shown on the right plot and is smaller than the number of input tokens (as expected). The two slowest models in terms of time to generate a token (LLaMa3-8B and Yi-6B) are among the fastest to complete the task. This can be explained by looking at the number of input and output tokens. The number of input tokens depends only on the tokenizer used by each model, because the input texts are the same for all models, the differences are small with LLaMa3-8B and Gemma using fewer tokens. Instead, the output texts depend on the responses of the models to the questions and there are large variations across models. The models that produce fewer tokens, are LLaMa3-8B and Yi-6B, so explaining the reasons for completing the tasks in less time than the other models. This clearly shows that the speed in generating tokens does not always correspond to the speed observed by the user.

The results for the second task are summarized in Figure 3. In this case, looking at the right plot, the number of tokens generated is rather similar to the number of input tokens as the models are just paraphrasing the input. The correlation between time per token and time to complete the task is better in this case, but there are still significant differences. For example, Mistral 7B is significantly faster in terms of time to complete the task than in generating tokens when compared to other models. This can be explained again by considering the number of tokens generated. The differences are still significant even for a task in which models are asked to paraphrase the input text.

Finally, the results for the third task are summarized in Figure 4. Again, the results in terms of token generation do not correspond with those of the time needed to complete the task. LLaMa-2-7B is fast in generating tokens, but among the slowest in completing the task. Instead, the Gemma model is the fastest because it generates substantially fewer tokens.

These results illustrate the complexity of evaluating the speed of LLMs (and thus also energy dissipation) under a fair comparison; even when considering the same hardware, the same prompts, and models of similar sizes, the time needed to complete a given task can be significantly different. Additionally, the relative performance of the models does not always correlate with their speed in generating tokens due to the use of different tokenizers and different models generate different number of tokens for the same prompt; moreover, the time to complete a task for the same model may be different depending on its parameters, such as temperature.

Therefore, speed metrics in terms of tokens per second or the time to generate a given number of tokens should be very carefully taken into consideration. A detailed evaluation per task is needed to truly understand the speed of different LLMs for a given scenario. A potential alternative is to develop speed benchmarks that are focused on specific tasks to complement existing per token metrics. For example, the definition of a set of input datasets and tasks (such as translation, summarization, question answering or essay writing) and the use of the time to complete the tasks must be utilized as part of the speed evaluation metrics in addition to the number of tokens per second. The availability of such benchmarks would enable a more comprehensive evaluation and comparison of the speed and energy dissipation of LLMs.

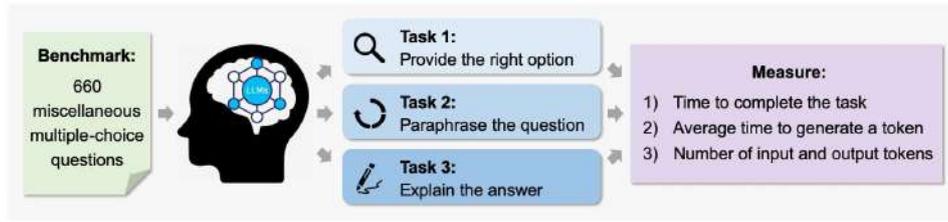

**FIGURE 1.** Overview of the evaluation procedure for the speed of an LLM.

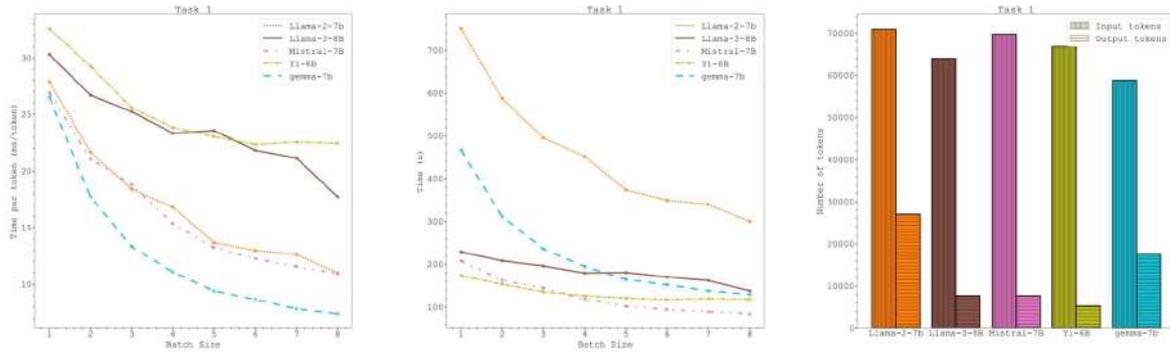

**FIGURE 2.** Results for task 1 answering 660 multiple choice questions: time to generate a token (left) time to complete the task (middle) and number of input and output tokens (right)

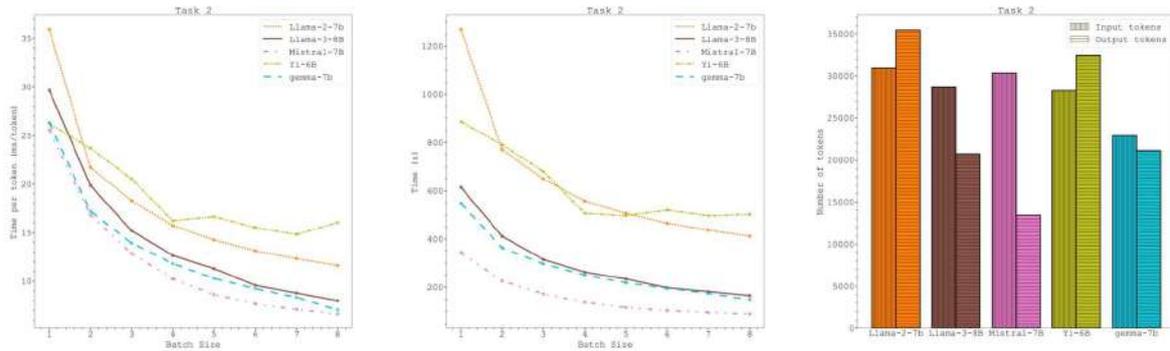

**FIGURE 3.** Results for task 2 paraphrasing 660 multiple choice questions: time to generate a token (left) time to complete the task (middle) and number of input and output tokens (right)

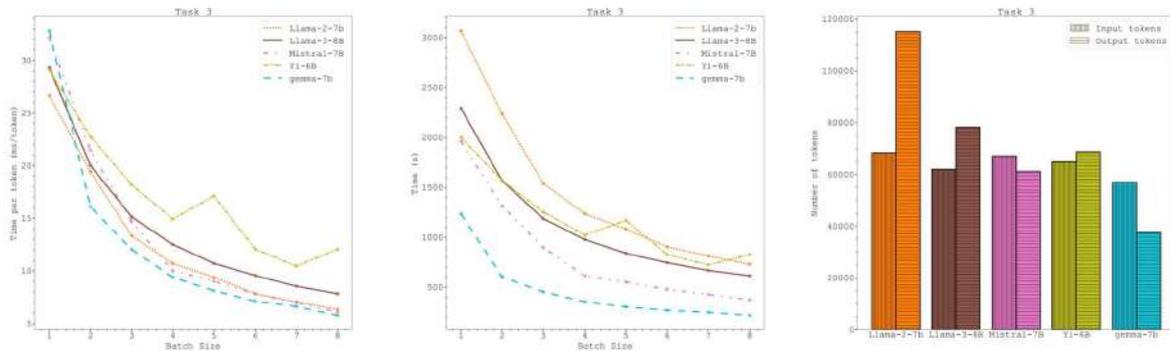

**FIGURE 4.** Results for task 3 open answers to 660 multiple choice questions: time to generate a token (left) time to complete the task (middle) and number of input and output tokens (right)

## CONCLUSION

As LLMs are widely used, their speed and energy dissipation have become a key issue. Existing benchmarks for assessing the speed of LLMs focus on the time needed to generate several tokens, or the number of tokens generated per second. However, the critical figures of merit are the time and energy needed to complete a given task. In this paper, we have studied the performance of several widely used open-weights LLMs when performing three simple tasks. The results show that the time needed to complete each task does not necessarily correlate to the number of tokens that the LLM can generate per second. This is due to the difference in the tokenizers, but more importantly on the lengths of the texts generated by each model for the same task. It also shows the complexity of evaluating LLM speed in realistic applications. To address this challenge, an alternative approach is to develop task-oriented benchmarks that are representative of LLM used cases. Such benchmarks could be more informative as to the relative speed of LLMs when performing a given task.

## ACKNOWLEDGMENTS

This work was supported by the Agencia Estatal de Investigación (AEI) (doi:10.13039/501100011033) under Grant FUN4DATE (PID2022-136684OB-C22), by the European Commission through the Chips Act Joint Undertaking project SMARTY (Grant no. 101140087) and by NVIDIA with a donation of GPUs.

**Javier Conde** is an assistant professor at the ETSI de Telecomunicación, Universidad Politécnica de Madrid, 28040 Madrid, Spain. Contact him at javier.conde.diaz@upm.es

**Miguel González** is a researcher at the ETSI de Telecomunicación, Universidad Politécnica de Madrid, 28040 Madrid, Spain. Contact him at miguel.gonsaiz@upm.es

**Pedro Reviriego** is an associate professor at the ETSI de Telecomunicación, Universidad Politécnica de Madrid, 28040 Madrid, Spain. Contact him at pedro.reviriego@upm.es

**Zhen Gao** is an associate professor at Tianjin University, 300072 Tianjin, China. Contact him at zgao@tju.edu.cn

**Shanshan Liu** is a professor at the University of Electronic Science and Technology of China, Chengdu 611731, Sichuan, China. Contact her at ssliu@uestc.edu.cn

**Fabrizio Lombardi** is a professor at Northeastern University, Boston, MA 02115, USA. Contact him at lombardi@coe.northeastern.edu